\RequirePackage[2020-02-02]{latexrelease}
\documentclass[times, twoside]{zHenriquesLab-StyleBioRxiv}
\usepackage{graphicx}
\usepackage[document]{ragged2e}

\leadauthor{Robinson} 

\begin{document}

\title{Assessing gender bias in medical and scientific masked language models with StereoSet}
\shorttitle{Gender Bias in Medical MLMs}

\author[1]{Robert Robinson, MD}

\affil[1]{Department of Internal Medicine, SIU School of Medicine, Springfield IL, USA}

\maketitle

\begin{abstract}
Background: NLP systems use language models such as Masked Language Models (MLMs) that are pre-trained on large quantities of text such as Wikipedia create representations of language.  BERT (Bidirectional Encoder Representations from Transformers) is a powerful and flexible general-purpose MLM system developed using unlabeled text.  The speed of development and accuracy make this approach attractive for many uses, including medical and scientific text analysis.  This type of pre-training on large quantities of text also transparently embeds the cultural and social biases found in the source text into the MLM system.  This study aims to compare biases in general purpose and medical MLMs with the StereoSet bias assessment tool.   
\hfill\break
\hfill\break
Methods: StereoSet was cloned from GitHub and installed in a Google Colab notebook with GPU support.  Huggingface transformers were used interface with the MLMs and obtain StereoSet scores for each MLM model tested. 
\hfill\break
\hfill\break
Results: The general purpose MLMs showed significant bias overall, with BERT scoring 57 and RoBERTa scoring 61.  The category of gender bias is where the best performances were found, with 63 for BERT and 73 for RoBERTa.  Performances for profession, race, and religion were similar to the overall bias scores for the general-purpose MLMs.  Medical MLMs showed more bias in all categories than the general-purpose MLMs except for SciBERT, which showed a race bias score of 55, which was superior to the race bias score of 53 for BERT.   More gender (Medical 54-58 vs. General 63-73) and religious (46-54 vs. 58) biases were found with medical MLMs. 
\hfill\break
\hfill\break
Conclusion:  This evaluation of four medical MLMs for stereotyped assessments about race, gender, religion, and profession showed inferior performance to general-purpose MLMs. These medically focused MLMs differ considerably in training source data, which is likely the root cause of the differences in ratings for stereotyped biases from the StereoSet tool.
\end {abstract}

\begin{keywords}
Gender bias | MLM | NLP | StereoSet
\end{keywords}

\begin{corrauthor}
rrobinson\at siumed.edu
\end{corrauthor}

\vfill\eject

\section*{Introduction}
Natural language processing (NLP) is a powerful tool for interacting with text content for facilitating automation, summarization, and analysis. The foundational components of many NLP systems are language models such as Masked Language Models (MLMs) that are pre-trained on large quantities of text such as Wikipedia (3.9 billion words by June 2021) or BooksCorpus (1 billion words from 11,038 English language books) to create representations of language [1, 2]. BERT (Bidirectional Encoder Representations from Transformers) is a powerful and flexible general-purpose MLM system developed by the Google AI Language team members using this approach [1]. \hfill\break
\hfill\break
The BERT development team employed bi-directional (right-to-left AND left-to-right) language encoding on a large volume of unlabeled text from Wikipedia and the BooksCorpus. The performance of BERT at many-core NLP tasks has been excellent [1], has resulted in over 28,000 citations by November of 2021 [3], and became a component of the Google search text analysis in 2019 [4]. The high performance of BERT and the ability to use unlabeled text for pre-training made this approach attractive for the development of specialist language MLMs.
 \hfill\break
\hfill\break
Specialized MLMs using the methods used to develop BERT were trained with specialty-specific sources of text such as SemanticScholar (full-text scientific journal articles) and the MIMIC III database (hospital discharge summaries) to produce medical and scientific specialty MLMs such as SciBERT [5], BioClinicalBERT [6], BioDischargeSummaryBERT [6], and CORe-clinical-outcome-BioBERT [7] became available. These efforts produced specialty language MLMs with high performance and accuracy for benchmark tasks with complex technical language. 
\hfill\break
\hfill\break
Pre-training MLMs on large quantities of text also transparently embeds the cultural and social biases found in the source text into the underlying system [8,9]. These embedded biases can influence search engine results, text automation, and text analysis in ways that may discriminate against people based on gender, race, culture, and other factors [10, 11]. Demonstrations of trained-in gender bias in NLP can be found for career roles and different situations. This embedded bias has the potential for implicit and explicit harm to people. 
\hfill\break
\hfill\break
To better investigate the scope and severity of biases in MLMs, tools such as StereoSet have been developed. 
 \hfill\break
\hfill\break
StereoSet quantifies stereotyped assessments made by MLMs about race, gender, religion, and profession using a crowd-sourced database of 17,000 test sentences. This tool uses a evaluation of the most probable word selected by a MLM to ‘fill in the blank’ in intra-sentence and inter-sentence Context Association Tests (CATs).  These most probable words to fill in the blank are then compared to human scored most probable words that are stereotypes, anti-stereotypes, or unrelated words in a given context.  This comparison is used to calculate a bias score ranging from high (score of 0) to low (score of 100) is calculated for each type of bias with this tool, facilitating comparisons between MLMs [12]. 
\hfill\break
\hfill\break
This study aims to compare biases in general purpose and medical MLMs with the StereoSet assessment tool. 
\hfill\break
\section*{Materials and Methods}
StereoSet was cloned from GitHub and installed in a Google Colab notebook with GPU support (https://github.com/moinnadeem/StereoSet).  Huggingface transformers were used interface with the MLMs. \hfill\break  
StereoSet scores were obtained for these MLMs:
\hfill\break
\hfill\break
\textbf{{BERT}}\hfill\break
BERT is an open source general-purpose MLM trained on BooksCorpus (800M words) and English Wikipedia (2,500M words) developed by the Google AI Language team [1].\hfill\break
Huggingface model card: \hfill\break
https://huggingface.co/bert-base-uncased\hfill\break 
May 18, 2021 Update
\hfill\break
\hfill\break
\textbf{{RoBERTa}}\hfill\break
RoBERTa is a general purpose open source MLM trained on BookCorpus, English Wikipedia, CC-News and OpenWebText developed by the Facebook AI team and other collaborators [2].\hfill\break
Huggingface model card: \hfill\break
https://huggingface.co/roberta-base\hfill\break
December 11, 2020 Update.
\hfill\break
\hfill\break
\textbf{{SciBert}}\hfill\break
SciBERT is an open source MLM for scientific text trained on 1.14 million full text journal articles from SemanticScholar developed by the Allen Institute for Artificial Intelligence [5].\hfill\break
Huggingface model card:  \hfill\break https://huggingface.co/allenai/scibert\_scivocab\_uncased
\hfill\break
May 19, 2021 Update.
\hfill\break
\hfill\break
\textbf{{BioClinicalBERT}} \hfill\break
BioClinicalBERT is an open source MLM for medical text trained on all notes from MIMIC III (880 million words) in addition to the BioBERT [14] by Alsentzer and colleagues [6].\hfill\break
Huggingface model card: \hfill\break https://huggingface.co/emilyalsentzer/Bio\_ClinicalBERT
\hfill\break
May 19, 2021 Update.
\hfill\break
\hfill\break
\textbf{{BioDischargeSummaryBERT}} \hfill\break
BioDischargeSummaryBERT is an open source MLM for medical text trained on all discharge summaries from MIMIC III (880 million words) by Alsentzer and colleagues [6].\hfill\break
Huggingface model card:\hfill\break https://huggingface.co/emilyalsentzer/Bio\_Discharge\_Summary\_BERT\hfill\break
May 19, 2021 Update.
\hfill\break
\hfill\break
\textbf{{COReclinicaloutcomeBioBERT}}\hfill\break
COReclinicaloutcomeBioBERT is an open source MLM for medical text builds on BioBERT by additional training MIMIC III, medical transcriptions from MTSamples, clinical notes from the i2b2 challenges, 10,000 case reports from PubMed Central (PMC), disease articles from Wikipedia and articles from MedQuAd.  This MLM optimized for clinical outcomes based NLP tasks was developed by an Aken and colleagues [7].\hfill\break
Huggingface model card: \hfill\break
https://huggingface.co/bvanaken/CORe-clinical-outcome-biobert-v1\hfill\break
May 19, 2021 Update.
\hfill\break
\hfill\break
\section*{Results}
 StereoSet bias scores range from 0 (high bias) to 100 (very low or no bias).  The general purpose MLMs showed significant bias overall, with BERT scoring 57 and RoBERTa scoring 61 (Table 1).  The category of gender bias is where the best performances were found, with 63 for BERT and 73 for RoBERTa. \hfill\break 
 \hfill\break
 Performances for profession, race, and religion were similar to the overall bias scores for the general-purpose MLMs.  The performance of RoBERTa was superior to BERT for overall bias, gender, profession, and race.  Religious bias was equal between BERT and RoBERTa.\hfill\break
 \hfill\break
The medical MLMs showed more bias in all categories than the general-purpose MLMs except for SciBERT, which showed a race bias score of 55, which was superior to the race bias score of 53 for BERT.   More gender (Medical 54-58 vs. General 63-73) and religious (46-54 vs. 58) biases were found with medical MLMs.  The CORe-clinical-outcome-BioBERT MLM showed a considerably worse racial bias score (46 vs. 53-58) than general-purpose MLMs.  \hfill\break
\hfill\break
The performance of CORe-clinical-outcome-BioBERT was worse than all tested general purpose and medical MLMs based on overall bias score, profession, and race scores. 
\hfill\break

\section*{Discussion}
This evaluation of four medical MLMs for stereotyped assessments about race, gender, religion, and profession showed inferior performance to general-purpose MLMs. 
These medically focused MLMs differ considerably in training source data, which is likely the root cause of the differences in ratings for stereotyped biases with the StereoSet tool because MLMs and other algorithmic language processing systems reflect the biases and stereotypes contained in the training text [10, 11, 13].  
BioBERT [14] and MIMIC III are training data for three out of the four medical MLMs tested [5, 6, 7].  SciBERT does not use these data sources and has superior performance overall and for every specific category of bias when assessed with StereoSet and is trained on the full texts of medical and scientific journal articles.\hfill\break
\hfill\break
MIMIC-III (Medical Information Mart for Intensive Care) is an extensive database comprising information relating to patients admitted to critical care units at Beth Israel Deaconess Medical Center in Boston, Massachusetts, USA, between 2001 and 2012.  Data, including clinical documentation, from over fifty thousand hospital admissions, are included in this dataset [15].  This clinical documentation from MIMIC III is a training source for BioClinicalBERT, BioDischargeSummaryBERT, and CORe-clinical-outcome-BioBERT.  \hfill\break\hfill\break
CORe-clinical-outcome-BioBERT expands clinical documentation to include medical transcriptions from MTSamples, clinical notes from the i2b2 challenges, and 10,000 case reports (clinical documentation like reports) from PubMed Central [7].  
\hfill\break\hfill\break
With these differences in training data noted, it is concerning that the medical MLMs that include training data from actual clinical documentation have more significant biases and stereotypes than found in general-purpose MLMs and SciBERT, which is trained only with journal article full texts.  This suggests that real-world clinical documentation from MIMIC III, MTSamples, and the i2b2 challenges may show more biased and stereotyped content than Wikipedia and BooksCorpus, which have been shown to include biased and stereotyped content [16, 17, 18, 19, 20].  
\hfill\break

\section*{Conclusions}
The potential impact of bias and discrimination in language models that may be used to suggest, inform, and evaluate text-based scientific and medical information is significant.  Further research is needed to clarify if these biases detected through the use of the StereoSet tool are due to the MLM methodology itself or if it reflects biases contained within the training data.

\section*{Bibliography}
\begin{itemize}
  
\item [1.]	Devlin, J., Chang, M.W., Lee, K. and Toutanova, K., 2018. Bert: Pre-training of deep bidirectional transformers for language understanding. arXiv preprint arXiv:1810.04805.
\hfill\break
\item [2.]	Liu, Y., Ott, M., Goyal, N., Du, J., Joshi, M., Chen, D., Levy, O., Lewis, M., Zettlemoyer, L. and Stoyanov, V., 2019. Roberta: A robustly optimized bert pretraining approach. arXiv preprint arXiv:1907.11692.
\hfill\break
\item [3.]	Google Scholar Search for “Bert: Pre-training of deep bidirectional transformers for language understanding".  https://bit.ly/3jMXwVx, Searched 10/31/2021
\hfill\break
\item [4.]	Schwartz, B.  2020.  Google: BERT now used on almost every English query. Search Engine Land. https://searchengineland.com/google-bert-used-on-almost-every-english-query-342193.
\hfill\break
\item [5.]	Beltagy I, Lo K, Cohan A. SciBERT: A pretrained language model for scientific text. arXiv preprint arXiv:1903.10676. 2019 Mar 26.
\hfill\break
\item [6.]	Alsentzer, E., Murphy, J.R., Boag, W., Weng, W.H., Jin, D., Naumann, T. and McDermott, M., 2019. Publicly available clinical BERT embeddings. arXiv preprint arXiv:1904.03323.
\hfill\break
\item [7.]	van Aken, B., Papaioannou, J.M., Mayrdorfer, M., Budde, K., Gers, F.A. and Löser, A., 2021. Clinical outcome prediction from admission notes using self-supervised knowledge integration. arXiv preprint arXiv:2102.04110.
\hfill\break
\item [8.]	Guo, W., Caliskan, A. (2020). Detecting emergent intersectional biases: Contextualized word embeddings contain a distribution of human-like biases. AAAI/ACM Conference on Artificial Intelligence, Ethics, and Society 2021. 
\hfill\break
\item [9.]	Kiritchenko, S. and Mohammad, S.M., 2018. Examining gender and race bias in two hundred sentiment analysis systems. arXiv preprint arXiv:1805.04508.
\hfill\break
\item [10.]	Bolukbasi, T., Chang, K.W., Zou, J., Saligrama, V. and Kalai, A., 2016. Man is to computer programmer as woman is to homemaker? debiasing word embeddings. arXiv preprint arXiv:1607.06520.
\hfill\break
\item [11.]	Manzini, T., Lim, Y.C., Tsvetkov, Y. and Black, A.W., 2019. Black is to criminal as caucasian is to police: Detecting and removing multiclass bias in word embeddings. arXiv preprint arXiv:1904.04047.
\hfill\break
\item [12.]	Nadeem, M., Bethke, A. and Reddy, S., 2020. Stereoset: Measuring stereotypical bias in pretrained language models. arXiv preprint arXiv:2004.09456.
\hfill\break
\item [13.]	Caliskan, A., Bryson, J.J. and Narayanan, A., 2017. Semantics derived automatically from language corpora contain human-like biases. Science, 356(6334), pp.183-186.
\hfill\break
\item [14.]	Lee, J., Yoon, W., Kim, S., Kim, D., Kim, S., So, C.H. and Kang, J., 2020. BioBERT: a pre-trained biomedical language representation model for biomedical text mining. Bioinformatics, 36(4), pp.1234-1240.
\hfill\break
\item [15.]	Johnson, A. E. W. et al. MIMIC-III, a freely accessible critical care database. Sci. Data 3:160035 doi: 10.1038/sdata.2016.35 (2016).
\hfill\break
\item [16.]	Schmahl, K.G., Viering, T.J., Makrodimitris, S., Jahfari, A.N., Tax, D. and Loog, M., 2020, November. Is Wikipedia succeeding in reducing gender bias? Assessing changes in gender bias in Wikipedia using word embeddings. In Proceedings of the Fourth Workshop on Natural Language Processing and Computational Social Science (pp. 94-103).
\hfill\break
\item [17.]	Hube, C., 2017, April. Bias in wikipedia. In Proceedings of the 26th International Conference on World Wide Web Companion (pp. 717-721).
\hfill\break
\item [18.]	Graells-Garrido, E., Lalmas, M. and Menczer, F., 2015, August. First women, second sex: Gender bias in Wikipedia. In Proceedings of the 26th ACM conference on hypertext and social media (pp. 165-174).
\hfill\break
\item [19.]	Callahan, E.S. and Herring, S.C., 2011. Cultural bias in Wikipedia content on famous persons. Journal of the American society for information science and technology, 62(10), pp.1899-1915.
\hfill\break
\item [20.]	Oeberst, A., von der Beck, I., Matschke, C., Ihme, T.A. and Cress, U., 2020. Collectively biased representations of the past: Ingroup Bias in Wikipedia articles about intergroup conflicts. British Journal of Social Psychology, 59(4), pp.791-818.
\hfill\break
\end{itemize}

\begin{contributions}
 
 Study design - RR\hfill\break
 Data acquisition - RR\hfill\break
 Data analysis - RR\hfill\break
 Writing, original and final draft - RR\hfill\break
\end{contributions}

\begin{interests}
 The author declares no competing financial interests.\hfill\break
 This study had no external funding.\hfill\break
\end{interests}
\newpage

\begin{figure}[ht!]
\centering
\caption{StereoSet Bias Scores of General Purpose and Medical MLMs}
\includegraphics[width=2\linewidth]{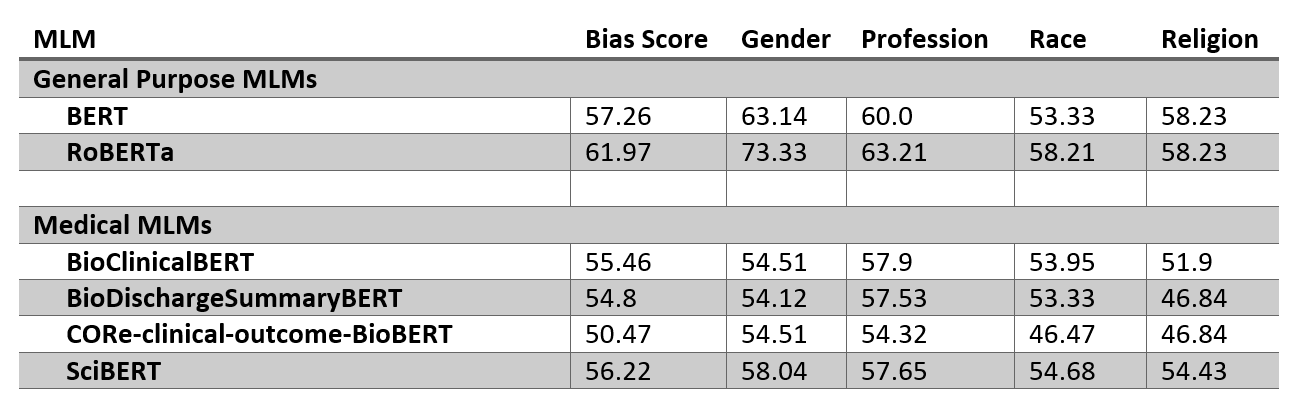}
\end{figure}

\end{document}